# Hybrid Fuzzy-ART based K-Means Clustering Methodology to Cellular Manufacturing Using Operational Time

Sourav Sengupta[1], Tamal Ghosh*[1], Pranab K Dan[1], Manojit Chattopadhyay[2]

[1]Department of Industrial Engineering & Management, West Bengal University of Technology

BF 142, Sector 1, Salt Lake City, Kolkata – 700064, West Bengal, India

sengupta.sourav86@gmail.com, tamal.31@gmail.com, danpk.wbut@gmail.com

[2]Department of Computer Application, Pailan College of Management & Technology

Bengal Pailan Park, 7000104, West Bengal, India

mjc02@rediffmail.com

*Abstract*— **This paper presents a new hybrid Fuzzy-ART based K-Means Clustering technique to solve the part machine grouping problem in cellular manufacturing systems considering operational time. The performance of the proposed technique is tested with problems from open literature and the results are compared to the existing clustering models such as simple K-means algorithm and modified ART1 algorithm using an efficient modified performance measure known as modified grouping efficiency (MGE) as found in the literature. The results support the better performance of the proposed algorithm. The Novelty of this study lies in the simple and efficient methodology to produce quick solutions for shop floor managers with least computational efforts and time.**

## I. INTRODUCTION

In response to the competitive markets need since the last three decades for increased industrial automation, product diversification and the trend towards shorter product life cycles, new manufacturing philosophies have been adopted by many of the established manufacturing firms. Among those new manufacturing philosophies, group technology (GT) has been used to reduce throughput and material handling times, to decrease work-in-progress and finished goods inventories and to increase the ability to handle forecast errors [1]. Group technology (GT) can be defined as a manufacturing philosophy identifying similar parts and grouping them together to take advantage of their similarities in manufacturing and design [2]. Cellular manufacturing (CM) is an application of GT and has emerged as a promising alternative manufacturing system. CM could be characterized as a hybrid system linking the advantages of both the jobbing (flexibility) and mass (efficient flow and high production rate) production approaches. CM entails the creation and operation of manufacturing cells. Parts are grouped into part families and machines into cells. As reported [3], the aim of CM is to reduce setup and flow times and therefore to reduce inventory and market response times. Setup times are reduced by using part-family tooling and sequencing, whereas flow times are reduced by minimizing setup and move times, wait times for moves and by using small transfer batches. Group Technology addresses issues such as average lot size decreasing, part variety increasing, increased variety of materials with diverse properties and requirements for closer tolerances. As described in a review [4], the basic idea behind GT/CM is to decompose a manufacturing system into subsystems by identifying and exploiting the similarities amongst part and machines. The very first step in this process is to solve the complex part machine grouping problem and the problem being quite challenging under real time scenario, various approaches have been developed, and among which soft computing approaches has an eminent role in the GT/CM literature. Soft Computing is the state-of-the-art approach to artificial intelligence which mostly comprises of Fuzzy Logic, Artificial Neural Network and Evolutionary Computing. This Paper presents a new hybrid neural network approach, Fuzzy-ART based K-Means Clustering Technique, to solve the part machine grouping problem in cellular manufacturing systems considering operation time. In light of the literature survey, it is well understood that very few studies focus on cell formation considering production factors such as operational time, operational sequence, batch size, production volume and other factors. In this work, it is attempted to form the cells considering operation time, a real time production factor. To solve such problem the zero-one MPIM of CF problem needs to be converted into real valued workload data. The workload represents the operational time required by the parts in the machines. The proposed model has been tested using wide variety of problems from literature and compared to the solutions obtained from simple k-means model and available modified ART1 model in the recent literature.

## II. LITERATURE REVIEW

Burbidge viewed group technology as a change from an organization of people mainly on process, to an organization based on completed products, components and major completed tasks [5]. Since 1960, various approaches were presented to solve the machine part grouping problem. Initially the methods like similarity coefficient methods (SCM) [6], graph theory [7] and rank order clustering (ROC) [8] methods were developed only to group the similar machines into machine cells while the grouping of parts into part families was done in the supplementary step of the procedure. Later clustering methods such as the MODROC [9] ,ZODIAC [10] MACE [11]are reported for solving the cell formation problems. Since late 80's soft-computing approaches began to gain popularity [4] and [12] which included artificial neural network , fuzzy logic and meta-heuristics like simulated annealing (SA) algorithm, genetic algorithm (GA), tabu search (TS).

*A. Overview of Artificial Neural Network*

Neural networks are massively parallel computer algorithms [13] with an ability to learn from experience. They have the capability to generalize, adapt, approximate given new information, and provide reliable classifications of data. These algorithms involve numerous computational nodes that have a high connectivity. Each of the nodes operates in a similar manner which makes them ideal for a parallel implementation. During the execution, each node receives an input, processes this information, and produces an output which is provided as an input to other nodes in the network. The connections between the nodes, and in particular the learning rules that modify the strength between the connections, give neural networks their power and flexibility [14].The neural network approach has been the subject of intensive study by interdisciplinary researchers for a long time. Though neural networks have been successfully applied in a variety of fields, their use in cellular manufacturing problems started in the late 80s and early 90s. Recognizing ANN"s pattern recognition ability, several researchers began to investigate neural network methods for the part-machine grouping problem. Neural networks are of major interest because when it is connected to computer, it mimics the brain and bombard people with much more information.

*B. Fuzzy Adaptive Resonance Theory*

Fuzzy ART proposed by Grossberg [15] belongs to the class of unsupervised, adaptive neural networks. Adaptive neural networks always had an important role in cellular manufacturing beginning in the early 90's in the works of [16], [17],[18] and [19]. Dagli and Huggahalli used ART1 in such problems while Malave and Ramachandran used competitive learning. Fuzzy ART was another common adaptive resonance framework as presented in the works of [20],[21], [22], [23], [24], [1] and [25] that provided a unified architecture for both binary and continuous valued inputs. Although Fuzzy ART does not require a completely binary representation of the parts to be grouped, it possesses the same desirable stability properties as ART1 and a simpler architecture than that of ART2. Figure 1 shows the architecture of the Fuzzy-ART network [26]. It consists of two layers of computing cells or neurons, and a vigilance subsystem controlled by an adjustable vigilance parameter. The input vectors are applied to the Fuzzy-ART network one by one. The network seeks for the "nearest" cluster that "resonates" with the input pattern according to a "winner-take-all" strategy and updates the cluster to become "closer" to the input vector. In the process, the vigilance parameter determines the similarity of the inputs belonging to the same cluster. For the same set of inputs, the similarity of elements in one cluster grows as the vigilance parameter increases, leading to a larger number of trained clusters. The choice parameter and the learning rate are two other factors that influence the quality of the clustering results. In this paper, Fuzzy ART is used to form the part families while k-means algorithm being an efficient clustering algorithm [27] is used to form the machine groups. The detailed description of the hybrid algorithm is discussed in the next section.

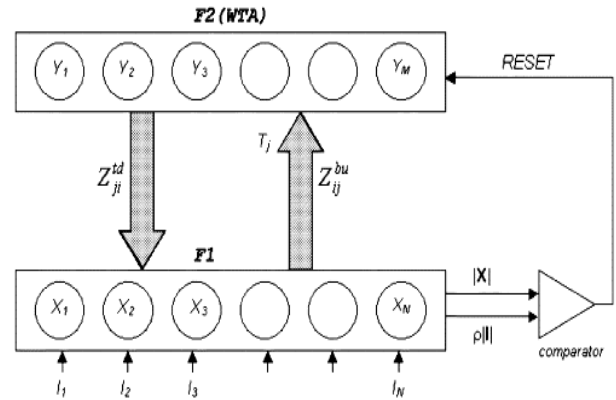

Fig. 1 Topological structure of the Fuzzy-ART architecture.

### III. THE PROPOSED HYBRID APPROACH

This study presents a hybrid Fuzzy-ART based K-Means Clustering Technique, a new pattern recognition neural network approach for clustering problems, and illustrates its use for machine cell design in group technology. The technique is a double mode clustering model. While mode1 is concerned with the identification of part families using the Fuzzy ART architecture, mode2 is concerned with the formation of machine groups using the K-Means clustering algorithm in order to obtain an informative and intelligent decision for the problem of designing a machine cell. The Fuzzy ART neural network was introduced by Carpenter et al. [15], and [28] implemented it to the CF problem. The latter found that in terms of bond energy recovery, Fuzzy ART outperformed ART1 and ART1/KS. The execution time of Fuzzy ART was higher than ART1 and ART1/KS, but for larger datasets, execution times were significantly lower than DCA and ROC2. The Fuzzy ART neural network involves several changes to ART 1: (a) non-binary input vectors can be processed; (b) there is a single weight vector connection ($w_{ij}$); and (c) in addition to vigilance threshold ($\rho$), two other parameters have to be specified: a choice parameter ($\alpha$) and a learning rate ($\beta$). The step-by-step illustration of Fuzzy ART network is as follows [28]:

Step1: Initialization
Connection weights: $w_{ij}(0)=1$.
$0 \leq I \leq N-1, 0 \leq j \leq (M-1)$
Select values for: $\alpha > 0, \beta \varepsilon (0, 1), \rho \varepsilon (0, 1)$
Step2: Read a new input vector I consisting of binary or analogue elements
Step3: Compute choice function ($T_j$) for every input node
$T_j = \| I \wedge w_j \| / [\alpha + \| w_j \|], 0 \leq j \leq (M-1)$,
Where $\wedge$ is the fuzzy AND operator, defined as : $(x \wedge y) = \min(x_i, y_i)$
Step4: Select the best matching exemplar .
$T_\Theta = \max \{ T_j \}$
Step5: Resonance test:
If $\| I \wedge w_\Theta \| / \| I \| \geq \rho$ then go to step 7 otherwise go to step 6
Step6: Mismatch reset: set $T_\Theta = -1$ and go to step 4
Step7: update best matching exemplar (learning law)
$w_\Theta^{new} = [\beta \times (I \wedge w_\Theta^{old})] + [(1 - \beta) \times w_\Theta^{old}]$
Step8: Repeat: go to step 2.

The above algorithm although could produce efficient clustering solutions, the literature suggests that hybrid approaches often produced better clusters. To establish the fact, K-Means clustering algorithm, a well known clustering technique with prominent results in the literature is integrated to the Fuzzy ART neural network to form the machine group based on the part families formed by the neural model.
K-Means [29] is a typical unsupervised clustering algorithm, which aims to partition N inputs (also called data points interchangeably) $x_1; x_2; \ldots ; x_N$ into $k^*$ clusters by assigning an input $x_t$ into the jth cluster if the indicator function $I(j\backslash X_t) = 1$ holds with

$$I(j/X_t) = \begin{cases} 1, & \text{if } j = \text{argmin}_{1 \leq r \leq k} \|X_t - m_r\|^2 \\ 0, & \text{otherwise} \end{cases} \quad (1)$$

Here, $m_1, m_2, \ldots, m_k$ are called seed points or units that can be learned in an adaptive way as follows:

**Step 1:** Pre-assign the number k of clusters, and initialize the seed points $\{m_j\}_{j=1}^k$
**Step 2:** Given an input Xt, calculate $I(j / X_t)$ by Eq.(1).
**Step 3:** Only update the winning seed point mw, i.e., $I(w/X_t) = 1$, by
$m_w^{new} = m_w^{old} + n(Xt - m_w^{old})$, (2)
where n is a small positive learning rate.
The above Step 2 and Step 3 are repeatedly implemented for each
input until all seed points converge.

In the hybrid model the K-Means algorithm is integrated into the Fuzzy ART architecture and on execution both the part and machine group clusters are produced. This integrated approach helps reduce computational time and often produce improved or comparable results when tested on the problems found in the literature. Another important aspect of the model proposed in this work is the ability to handle work load data. The model is equally capable of handling workload matrix as compared to part machine incidence matrix (MPIM) and the machine cells are formed based on the operation time, a real time production factor. The detailed step by step approach of the integrated model is presented below.

**Step 1: Input the workload matrix**
Machines in rows and parts in columns
**Step 2: Normalize input matrix by complement coding**
Step 2.1: Determine the size of the data.
    [totalNumofMachines, TotalNumofParts] = size(workloadMatrix);

Step 2.2: Create the return variable.
    C = ones(2* totalNumofMachines, TotalNumofParts);

Step 2.3: For each part do the complement coding
for j = 1: TotalNumofParts
  count = 1;
  for i = 1:2:(2* totalNumofMachines)
    C(i, j) = data(count, j);
    complementCodedData(i + 1, j) = 1 - data(count, j);
    count = count + 1;

**Step 3: Create and initialize the Fuzzy ART K-Means network**

Step 3.1: Create and initialize the weight matrix.
weight = ones(totalNumofMachines, 0);

Step 3.2: select the number of clusters 'K' for machine grouping using K-Means

Step 3.3: initial value of centroid for K-Means

Step 3.4: Create the structure and return
FuzzyArt = struct(' totalNumofMachines ',
{ totalNumofMachines }, 'TotalNumofCategories', {0},
'MaximumNumofCategories', {100}, 'weight', {weight}, ...
        'vigilance', {0.75}, 'bias', {0.000001},
'totalNumOfEpochs', {200}, 'learningRate', {1.0});

**Step 4: Training the Fuzzy ART network**

   Step 4.1: Set the return variables

    FuzzyArt = {};
    Classification = ones(1, TotalNumofParts);

   Step 4.2: for each epoch go through the complement coded workload matrix

   Step 4.3: Classify and learn on each part
   Step 4.3.1:  Activate the classifications
   Step 4.3.2: Rank the activations

Step 4.3.3: In the sorted list go through each classification and find the best match.

Step 4.3.4: must create a new classification if no classifications yet found

Step 4.3.5: Calculate the match
Step 4.3.6: if the match is greater than the vigilance then update the weights and induce resonance
Step 4.3.7: else choose the next classification in the sorted classification list
Step 4.4: if at the last epoch the network does not change at all, equilibrium is reached and stop training
**Step 5: Final Part machine clustering**
Step 5.1: Set up the return variables.
Classification = ones(1, TotalNumofParts );
Step 5.2: Classify and learn on each part
Step 5.3.1: Activate the classifications
Step 5.3.2: Rank the activations
Step 5.3.3: look for the best match
Step 5.4: if the match is greater than the vigilance then induce resonance
Step 5.5: else choose the next classification in the sorted classification list
If it is the last classification in the list, set the classification for the return value as -1 and induce resonance.
Step 5.6: from the return variable part group is identified
Step 5.6: check for the distance using K-Means for machine grouping
Step 5.7 : check for the minimum distance
Step 5.8: based on the return variable machine groups are identified.

## IV. RESULTS AND DISCUSSION

In this study, an efficient artificial neural network based hybrid model has been proposed for cell formation problem considering operational time of the parts instead of conventional zero-one incidence matrix with the objective of minimizing exceptional elements and voids while improving the grouping efficiency. In order to measure the grouping efficiency of an algorithm for machine-part CF, a performance measure is needed. Many performance measures for evaluating the goodness of PMG have been proposed over the years. Some popular performance measures that have been widely adopted in literature are grouping efficiency proposed by Chandrasekharan and Rajagopalan in 1986, grouping efficacy, proposed by Kumar and Chandrasekharan in 1990 and grouping capability index (GCI), proposed by Seifoddini & Hsu in 1994. However, the above-mentioned measures are not applicable to the proposed FAKMCT model for part machine grouping since they are based on the block diagonal configuration of binary PMIM and they do not incorporate the real-field data such as the operation time. To measure the clustering efficiency considering operation time, in this work modified grouping efficiency, [31] is used. The MGE is calculated using the following formulation.

$$MGE = \frac{T_{pti}}{T_{pto} + \sum_{k=1}^{c} T_{ptk} + \sum_{k=1}^{c} T_{ptk} \cdot \frac{N_{vk}}{N_{ek}}}$$

$T_{pti}$ : Total processing time inside the cells
$T_{pto}$ : Total processing time outside the cells
$T_{ptk}$ : Total processing time of cell k
$N_{vk}$ : No of voids in cell k
$N_{ek}$ : Total number of elements in cell k

Unlike grouping efficiency, modified grouping efficiency does not treat all the operations equally. Moreover a weighting factor for voids is considered to reflect the packing density of the cells. It produces 100% efficiency when the cells are perfectly packed without any voids and exceptional elements. The proposed algorithm is coded in Matlab7 and tested on the Intel Celeron M processor. The real valued workload matrix is presented to the algorithm as input. The proposed approach is tested on 10 well known datasets available in the GT/CM literature which were converted to workload matrix. The results obtained are compared to simple K-Means, and modified ART1 algorithm as presented in the work of [30].In order to compare the performance with the mentioned work in the literature, instead of generating the workload matrix from the PMIM in the literature in a random manner, the same workload matrix is taken as referred in the published work of [30].Around 50% of the solutions indicated improvement as measured in terms of minimum exceptional elements and maximum MGE. The learning rate is initialized to 1 and the vigilance parameter is considered as 0.75. Table I presents the source of the datasets used from the literature to demonstrate the proposed model and Table II presents the comparison between the results obtained from the proposed technique and the K-Means and modified ART1 available in literature.

TABLE I
SOURCE OF THE DATASETS USED FROM THE LITERATURE

| *Dataset No.* | *Sources* | *Problem size* |
|---|---|---|
| 1 | King and Nakornchai (1982) | 5 × 7 |
| 2 | Seiffodini (1989) | 5 × 18 |
| 3 | Kusiak (1987) | 7 × 11 |
| 4 | Seiffodini and Wolfe (1986) | 8 × 12 |
| 5 | Chandrasekharan et al. (1986)a | 8 × 20 |
| 6 | Mosier et al. (1985) | 10 × 10 |
| 7 | Askin et al. (1987) | 14 × 23 |
| 8 | Srinivasan et al. (1990) | 16 × 30 |
| 9 | Chandrasekharan et al. (1989)a | 24 × 40 |
| 10 | Stanfel (1985)a | 30 × 50 |

From Table I and Table II, it could be seen that a wide variety of datasets have been chosen from the literature with part machine workload matrices ranging from 5×7 to 30×50. Most of the solutions obtained from the proposed hybrid neural approach outperformed the other two while the rest demonstrated similar results. In dataset 5 and dataset 7 the count of exceptional elements reduced by 3 and 2.In dataset 8 the result observed shows an improvement of 2.39% MGE with just 1 exceptional added to the solution set while datasets 14 and dataset 3 showed clear improvements in terms of exceptional elements and MGE.

TABLE II
COMPARISON AMONG K-MEANS, MODIFIED ART1 AND HYBRID APPROACH

| DS | NOC | K-Means | | M-ART1 | | HYBRID | |
|---|---|---|---|---|---|---|---|
|  |  | EE | MGE | EE | MGE | EE | MGE |
| 1 | 2 | 2 | 77.25 | 2 | 77.25 | **2** | **77.26** |
| 2 | 2 | 7 | 81.87 | 7 | 81.87 | **7** | **81.88** |
| 3 | 2 | 3 | 61.77 | 3 | 61.77 | **3** | **62.06*** |
| 4 | 2 | 6 | 57 | 4 | 69.7 | **4** | **64.15** |
| 5 | 2 | 28 | 60 | 25 | 61.3 | **22** | **60.75*** |
| 6 | 3 | 0 | 77.14 | 0 | 77.14 | **0** | **77.17** |
| 7 | 2 | 2 | 59.43 | 2 | 60.59 | **0** | **59.81*** |
| 8 | 3 | 15 | 64.81 | 15 | 64.81 | **16** | **67.2*** |
| 9 | 6 | 0 | 90.28 | 0 | 90.28 | **0** | **90.28** |
| 10 | 6 | 20 | 61.84 | 26 | 55.51 | **17** | **59.3*** |

*: improvement; EE: exceptional elements; M-ART1: modified ART1; NOC: number of cells formed; MGE: modified grouping efficiency

Thus from the results the efficiency of the model in handling ratio level data and capability of clustering machine part workload matrix with minimum exceptional elements and maximum possible MGE is justified and hence could be established as a simple and efficient methodology to produce quick solutions for shop floor managers with least computational efforts and time. Fig. 2 shows an input workload matrix while Fig. 3 and Fig. 4 demonstrate the solution sets obtained from the proposed technique.

|    | p1   | p2   | p3   | p4   | p5   | p6   | p7   | p8   | p9   | p10  | p11  |
|----|------|------|------|------|------|------|------|------|------|------|------|
| m1 | 0    | 0.53 | 0.99 | 0    | 0    | 0    | 0.83 | 0    | 0    | 0    | 0    |
| m2 | 0.91 | 0    | 0    | 0    | 0.82 | 0    | 0    | 0    | 0    | 0    | 0.83 |
| m3 | 0    | 0    | 0    | 0    | 0    | 0    | 0    | 0    | 0.91 | 0.92 |      |
| m4 | 0.86 | 0    | 0.97 | 0    | 0    | 0.79 | 0    | 0    | 0    | 0    | 0    |
| m5 | 0    | 0    | 0    | 0    | 0.56 | 0    | 0    | 0.88 | 0    | 0    | 0    |
| m6 | 0.53 | 0    | 0    | 0.51 | 0    | 0    | 0    | 0.98 | 0.83 | 0.71 | 0    |
| m7 | 0    | 0    | 0.58 | 0.54 | 0    | 0.54 | 0.74 | 0    | 0.63 | 0    | 0    |

Fig. 2 Input workload Matrix for example problem of size 7 x 11, dataset 3

|    | p2   | p4   | p3   | p6   | p7   | p5   | p11  | p8   | p9   | p10  | p1   |
|----|------|------|------|------|------|------|------|------|------|------|------|
| m1 | 0.53 | 0    | 0.99 | 0    | 0.83 | 0    | 0    | 0    | 0    | 0    | 0    |
| m4 | 0    | 0    | 0.97 | 0.79 | 0    | 0    | 0    | 0    | 0    | 0    | 0.86 |
| m7 | 0    | 0.54 | 0.58 | 0.54 | 0.74 | 0    | 0    | 0    | 0.63 | 0    | 0    |
| m2 | 0    | 0    | 0    | 0    | 0    | 0.82 | 0.83 | 0    | 0    | 0    | 0.91 |
| m3 | 0    | 0    | 0    | 0    | 0    | 0    | 0.92 | 0    | 0    | 0.91 | 0    |
| m5 | 0    | 0    | 0    | 0    | 0    | 0.56 | 0    | 0.88 | 0    | 0    | 0    |
| m6 | 0    | 0.51 | 0    | 0    | 0    | 0    | 0    | 0.98 | 0.83 | 0.71 | 0.53 |

Fig. 3 Output matrix by the proposed FAKMCT based algorithm for example dataset 3

Fig. 4 Output matrix by the proposed neural based algorithm for example problem dataset 10

In cellular manufacturing systems the number of cells formed often has an effective role in maximizing MGE and minimizing exceptional elements. More number of cells increases exceptional elements thus affecting the MGE while in other cases it may increase MGE by reducing voids. Again under some circumstances it increases MGE without affecting the exceptional elements and could be referred as ideal number of cells. The optimal number of cells is thus required to find the best solution. The analysis is demonstrated in Fig. 5, 6, 7 and finally concluded in Fig. 8.

Fig. 5 shows a clustered result of a workload matrix of size 8×12 (dataset 4). In this case, 2 cells are formed with 4 exceptional elements and 64.15% MGE.

|    | p1 | p2 | p11 | p12 | p3 | p4 | p5 | p6 | p7 | p10 | p8 | p9 |
|----|----|----|-----|-----|----|----|----|----|----|-----|----|----|
| m1 | 0.53 | 0.99 | 0 | 0 | 0.83 | 0.91 | 0 | 0 | 0 | 0 | 0 | 0 |
| m7 | 0 | 0 | 0.68 | 0.67 | 0 | 0 | 0 | 0 | 0 | 0 | 0 | 0 |
| m8 | 0 | 0 | 0.7 | 0.84 | 0 | 0 | 0 | 0 | 0 | 0 | 0 | 0 |
| m5 | 0 | 0 | 0 | 0 | 0 | 0 | 0 | 0 | 0.63 | 0.63 | 0.53 | 0.69 |
| m6 | 0 | 0 | 0.94 | 0 | 0 | 0 | 0 | 0 | 0.68 | 0 | 0.51 | 0.61 |
| m2 | 0.82 | 0 | 0 | 0 | 0.83 | 0.91 | 0.92 | 0.86 | 0.97 | 0.79 | 0 | 0 |
| m3 | 0 | 0 | 0 | 0 | 0.56 | 0.88 | 0.53 | 0.51 | 0.98 | 0 | 0.83 | 0.71 |
| m4 | 0 | 0 | 0 | 0 | 0 | 0 | 0 | 0.58 | 0.54 | 0.63 | 0.54 | 0.74 |

Fig. 5 Dataset 4 with 2 cells

Fig. 6 shows the clustered result of the same workload matrix of size 8×12 (dataset 4) as in Fig. 5. In this case, 3 cells are formed. Number of exceptional elements is still 4 while MGE changes to 68.56%.

|    | p1 | p2 | p11 | p12 | p3 | p4 | p5 | p6 | p7 | p10 | p8 | p9 |
|----|----|----|-----|-----|----|----|----|----|----|-----|----|----|
| m1 | 0.53 | 0.99 | 0 | 0 | 0.83 | 0.91 | 0 | 0 | 0 | 0 | 0 | 0 |
| m7 | 0 | 0 | 0.68 | 0.67 | 0 | 0 | 0 | 0 | 0 | 0 | 0 | 0 |
| m8 | 0 | 0 | 0.7 | 0.84 | 0 | 0 | 0 | 0 | 0 | 0 | 0 | 0 |
| m5 | 0 | 0 | 0 | 0 | 0 | 0 | 0 | 0 | 0.63 | 0.63 | 0.53 | 0.69 |
| m6 | 0 | 0 | 0.94 | 0 | 0 | 0 | 0 | 0 | 0.68 | 0 | 0.51 | 0.61 |
| m2 | 0.82 | 0 | 0 | 0 | 0.83 | 0.91 | 0.92 | 0.86 | 0.97 | 0.79 | 0 | 0 |
| m3 | 0 | 0 | 0 | 0 | 0.56 | 0.88 | 0.53 | 0.51 | 0.98 | 0 | 0.83 | 0.71 |
| m4 | 0 | 0 | 0 | 0 | 0 | 0 | 0 | 0.58 | 0.54 | 0.63 | 0.54 | 0.74 |

Fig. 6 Dataset 4 with 3 cells

Fig. 7 also shows the clustered result of the same workload matrix of size 8×12 (dataset 4) as in Fig. 5 and Fig. 6. In this case, 4 cells are formed. Number of exceptional elements increases to 10 while MGE reduces to 65.86%.

|    | p1 | p2 | p11 | p12 | p3 | p4 | p5 | p6 | p7 | p10 | p8 | p9 |
|----|----|----|-----|-----|----|----|----|----|----|-----|----|----|
| m1 | 0.53 | 0.99 | 0 | 0 | 0.83 | 0.91 | 0 | 0 | 0 | 0 | 0 | 0 |
| m7 | 0 | 0 | 0.68 | 0.67 | 0 | 0 | 0 | 0 | 0 | 0 | 0 | 0 |
| m8 | 0 | 0 | 0.7 | 0.84 | 0 | 0 | 0 | 0 | 0 | 0 | 0 | 0 |
| m2 | 0.82 | 0 | 0 | 0 | 0.83 | 0.91 | 0.92 | 0.86 | 0.97 | 0.79 | 0 | 0 |
| m3 | 0 | 0 | 0 | 0 | 0.56 | 0.88 | 0.53 | 0.51 | 0.98 | 0 | 0.83 | 0.71 |
| m5 | 0 | 0 | 0 | 0 | 0 | 0 | 0 | 0 | 0.63 | 0.63 | 0.53 | 0.69 |
| m6 | 0 | 0 | 0.94 | 0 | 0 | 0 | 0 | 0 | 0.68 | 0 | 0.51 | 0.61 |
| m4 | 0 | 0 | 0 | 0 | 0 | 0 | 0 | 0.58 | 0.54 | 0.63 | 0.54 | 0.74 |

Fig. 7 Dataset 4 with 4 cells

From the above experiment it could be seen that for the referred workload matrix of size 8×12, the optimal number of cells needed is 3 which gives the best solution as shown in Fig. 8.

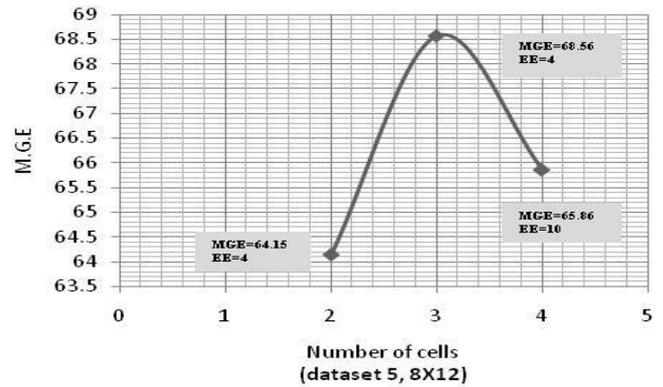

Fig. 8 Optimal number of cells, dataset 4 (8×12)

## V. CONCLUSION

In this work an artificial neural network based hybrid clustering model is proposed to solve the cell formation problem using the non binary real valued work load data as an input matrix. The proposed algorithm is tested with benchmark problems found in the literature and the results are compared with the existing algorithms mainly K-means clustering algorithm and the modified ART1 algorithm. To measure the performance of the proposed model considering the ratio level data modified grouping efficiency (MGE) is used. The results obtained which often outperformed the results in the literature justified the efficiency of the model in cell formation and sets a new milestone for the hybrid neural network approaches in GT/CM literature. The work can be further extended in future incorporating production data like operation sequence, machine capacity, production volume, layout considerations and material handling systems which further enhance the problem into a more generalized form in manufacturing environment.